\documentclass[submission,copyright,creativecommons, noncommercial]{eptcs}
\providecommand{\event}{FMAS 2023}

\usepackage{rotating}
\usepackage{graphicx}
\usepackage[T1]{fontenc}
\usepackage[utf8]{inputenc}
\usepackage{cite}
\usepackage{amsmath,amssymb,amsfonts}
\usepackage{algorithmic}
\usepackage{graphicx}
\usepackage[bottom]{footmisc}
\usepackage{hyperref}
\usepackage{csquotes}
\usepackage{textcomp}
\usepackage{enumitem,kantlipsum}
\usepackage{xcolor}
\usepackage{xspace}
\usepackage{multirow}
\def\BibTeX{{\rm B\kern-.05em{\sc i\kern-.025em b}\kern-.08em
    T\kern-.1667em\lower.7ex\hbox{E}\kern-.125emX}}

\usepackage{graphicx}
\usepackage{indentfirst}
\begin{document}

\pagenumbering{arabic} 
\pagestyle{plain}

\title{Robustness Verification for Knowledge-Based \\ Logic of Risky Driving Scenes}
%
%
\author{
\and
\quad Xia Wang
\institute{\quad Vanderbilt University \\ \quad Nashville, TN, USA}
\email{\quad xia.wang@vanderbilt.edu}
\and
\quad\quad\quad Anda Liang
\institute{\quad\quad\quad\quad Vanderbilt University \\ \quad\quad\quad\quad Nashville, TN, USA}
\email{\quad\quad\quad\quad anda.liang@vanderbilt.edu}
\and
\and
Jonathan Sprinkle
\institute{Vanderbilt University \\ Nashville, TN, USA}
\email{johnathan.sprinkle@vanderbilt.edu}
\and
Taylor T. Johnson
\institute{Vanderbilt University \\ Nashville, TN, USA}
\email{taylor.johnson@vanderbilt.edu}
}

\def\titlerunning{Robustness Verification for Risky Driving Scenes}
\def\authorrunning{X. Wang, A. Liang, J. Sprinkle, \& T.T. Johnson}

\maketitle 

\begin{abstract}
\textbf{Abstract.} Many decision-making scenarios in modern life benefit from the decision support of artificial intelligence algorithms, which focus on a data-driven philosophy and automated programs or systems. However, crucial decision issues related to security, fairness, and privacy should consider more human knowledge and principles to supervise such AI algorithms to reach more proper solutions and to benefit society more effectively. 
In this work, we extract knowledge-based logic that defines risky driving formats learned from public transportation accident datasets, which haven't been analyzed in detail to the best of our knowledge. More importantly, this knowledge is critical for recognizing traffic hazards and could supervise and improve AI models in safety-critical systems. Then we use automated verification methods to verify the robustness of such logic.
More specifically, we gather 72 
accident 
datasets from Data.gov\footnote{\url{https://catalog.data.gov/dataset/?tags=crash}} and organize them by state. Further, we train Decision Tree and XGBoost models on each state's dataset, deriving accident judgment logic. Finally, we deploy robustness verification on these tree-based models under multiple parameter combinations.

\vspace{0.5cm}\noindent

\end{abstract}

\section{Introduction}
As real-world domains become more complex, sophisticated, and dynamic, knowledge-based systems face the challenge of preserving the coherence and soundness of their knowledge bases. 
Moreover, since the current mainstream autonomous driving technology still requires the intervention of human drivers, the interpretability of such driving assistance technologies and algorithms, as well as the use of human-understandable logic to build driving scenario applications, has become an essential requirement.
Finally, since the application scenario of such data-driven and human-understandable logic is closely related to security issues, it is becoming increasingly important for the system knowledge base to be formally verified \cite{liu1996formal}.
Understanding and learning risky driving properties via corresponding traffic accident videos is an intuitive way.
Although it may be possible to train a neural network and directly output for the importance of each video frame, such an approach would require a large amount of hand-annotated data \cite{kumar2020video}. 
Directly collecting car accident videos from dashboard cameras would also be unethical. 
Additionally, only using deep learning models for video analysis is purely data-driven and therefore lacks comprehensiveness when the data availability is limited \cite{seshia2022toward}. 
To solve such issues mentioned above facing these obstacles, we propose to gather knowledge or rule logic for detecting risky driving behaviors from the public transportation accident dataset, which contains great value to assist other tasks such as anomaly detection and accident prevention but lacks of comprehensive analysis for now.


Since the data formats of different states are different
we train decision tree models and decision tree ensembles
on each state's dataset separately. Also, we employed a pre-established formal verification method\footnote{\href{https://github.com/chenhongge/treeVerification\#configuration-file-parameters}{https://github.com/chenhongge/treeVerification\#configuration-file-parameters}} \cite{chen2019robustness} to measure the robustness of these decision-based models. Such formal verification not only enhances our models but is also necessary.
In fact, recent studies have demonstrated that neural network models are vulnerable to adversarial perturbations—a small and human imperceptible input perturbation can easily change the predicted label \cite{szegedy2013intriguing, carlini2017towards, goodfellow2014explaining}. This has created serious security threats to many real applications so it becomes important to formally verify the robustness of machine learning models. Usually, the robustness verification problem can be cast as finding the minimal adversarial perturbation to an input example that can change the predicted class label. In this work, we build tree-based models to gain human-understandable logic to support driving and transportation management tasks. Thus, we target the studies focusing on robustness verification of tree-based models. Recent studies have demonstrated decision tree models are also vulnerable to adversarial perturbations \cite{chen2019robust, cheng2018query, kantchelian2016evasion}. 

Tree-based models may provide additional insights into situations that result in accidents, which may have benefits for safety and could save lives. However, it is important to understand their robustness in making accurate predictions.
In this paper, we will describe our approach to constructing tree-based models using publicly available accident datasets to determine the characteristics of risky driving scenarios and verify the robustness of those decision trees. To summarize, our framework offers the following contributions:
\begin{itemize}[leftmargin=*,labelindent=0pt,noitemsep,topsep=3pt]
    \item To the best of our knowledge, this work is the first comprehensive collation and analysis of large-scale traffic accident data across the United States.
    \item We extract human-understandable rule and logic to further support driving and transportation management tasks, and even safety-critical AI tasks related to traffic scenarios.
    \item We provide suggestions on unified accident data collection, which could be a guidance for different states to follow and to gather unified data format of traffic accident recordings.
\end{itemize}

\section{Related Work}
Formal verification of risky driving situations has been an active area of research for the past few years. This section summarizes some of the relevant works in this field, organized into two subsections: modeling of driving scenarios and formal verification techniques.

\subsection{Modeling of Driving Scenarios}
There have been many research efforts focused on the development of realistic models for driving scenarios, which are used as inputs for formal verification tools. For example, prior research has developed stochastic models of drivers' behavior based on Reachability Analysis \cite{tran2019safety}, Convex Markov Chains \cite{sadigh2014data} and tools such as the CARLA simulator, for validation of autonomous driving systems \cite{dosovitskiy2017carla}.  We seek to explore different models and simulators with public transportation datasets and apply formal verification to check the logic and/or rule for consistency and comprehensiveness.

Our proposed method combines insights from autonomous vehicle studies and formal verification literature. Alawadhi et al.'s review work on autonomous vehicle adoption factors, including safety, liability, and trust, forms a crucial base \cite{alawadhi2020systematic}. Many researchers, like Tahir and Alexander, have proposed coverage-based testing for self-driving vehicles. Their paper aims to increase public confidence in self-driving autonomous vehicles by verifying and validating the techniques used in their development \cite{tahir2020coverage}.


Other research has tackled the challenge of ensuring the coherence and soundness of knowledge-based systems, including those used for video processing. For example, Liu et al. proposed a formal verification method for knowledge-based systems using Petri networks to analyze the reachability of certain states \cite{liu1996formal}. Meanwhile, Kumar et al. explored the application of deep learning models for video analysis \cite{kumar2020video}. Although this study does not employ formal verification methods, it provides insights into the coordination and protocols needed to anticipate, predict, and prevent accidents at common locations such as intersections. Other studies have also examined the general aspects of knowledge-based systems, demonstrating the versatility of formal verification methods in this area.

\subsection{Formal Verification Techniques}
Many formal verification techniques have been proposed for the verification of driving scenarios. These techniques range from model checking and theorem proving to more advanced methods such as abstraction and constraint-based analysis. In fact, prior studies have examined a scenario-based approach for formal modeling \cite{xu2019scenario} and scenario-based probabilistic collision risk estimator \cite{ledent2019formal}. 

Robustness verification is a process of evaluating the resilience of a system or a model to different types of perturbations or uncertainties. This type of verification is crucial for ensuring the reliability and safety of complex systems, such as autonomous vehicles, aerospace systems, and medical devices. As we mentioned above, there are several works that focus on robustness verification of deep neural networks \cite{li2020prodeep,yang2021enhancing}. Björn et al. \cite{lutjens2020certified} propose an approach to verify the robustness of the reinforcement learning algorithm, which is demonstrated on a Deep Q-Network policy and is shown to increase robustness to noise and adversaries in pedestrian collision avoidance scenarios and a classic control task. Zhouxing et al. \cite{shi2020robustness} provides a formal robustness verification method for Transformers that have complex self-attention layers. Also, some works investigate applications in safety-critical domains, such as autonomous vehicles \cite{zhang2022robustness, sadigh2019verifying}. 

In this work, we utilize the robustness verification method on tree-based models \cite{chen2019robustness}\footnote{\url{https://github.com/chenhongge/treeVerification}} to verify the robustness of the accident detection logic gained from government crash datasets.

\section{Proposed Method}

\subsection{System Overview}

\begin{figure}[ht]
     \centering
     \includegraphics[width=.85\textwidth]{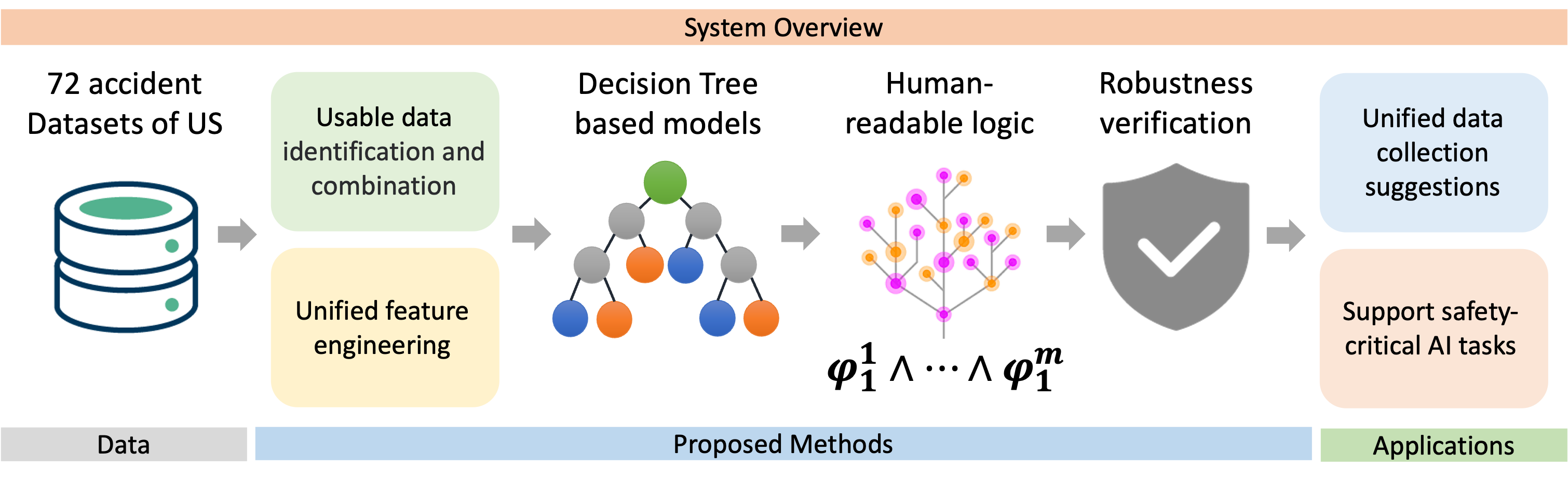}
     \caption{Our approach leverages nationwide accident datasets, tree-based models, and robustness verification technologies. The main contributions of this work are 1) it provides Unified data collection suggestions, and 2) it provides human-readable logic that could support safety-critical AI tasks.}
     \label{fig:robust_overview}
 \end{figure}

Fig.~\ref{fig:robust_overview} depicts an overview of our proposed approach. 

\subsection{Unified Feature Engineering}

The unified feature engineering step is an essential part of the data analysis process, as it ensures that the data is uniform and can be used to generate accurate results. 
In different states, the category values of one common feature vary due to different encoding policies in different states or errors in manual recording, so the key data pre-processing progress is to standardize and unify the encoding policy for these features. 

For example, the category values of collision types in each state are quite different. 
We refer to the unified manner of collision code\footnote{\label{note1}\url{https://masscrashreportmanual.com/crash/manner-of-collision/}}, which includes 11 collision categories in total, to unify the category values of Maryland and Arizona shown in Table \ref{tab:Collisionmanner}. The detailed code for all unified feature engineering processes can be found in this repository: \url{https://github.com/WilliamStar007/decision-tree}.

\begin{table}[h!]
\centering
\caption{Unified collision manner values mapping policy of Maryland and Arizona.}
\resizebox{\columnwidth}{!}{%
\begin{tabular}{lllll}
\hline
\textbf{Code  \;} &
  \textbf{Attribute} &
  \textbf{Definition} &
  \textbf{Maryland State mapping values} &
  \textbf{Arizona State mapping values} \\ \hline
1 &
  \begin{tabular}[c]{@{}l@{}}Single\\ Vehicle\\ Crash\end{tabular} &
  \begin{tabular}[c]{@{}l@{}}Indicates a crash involving no more \\ than one motor vehicle. Common \\ types of single-vehicle crashes are \\ noncollisions or crashes involving \\ pedestrians, fixed objects, wild animals \\ or unrestrained domestic animals.\end{tabular} &
  'Single   Vehicle' &
  'Single Vehicle' \\ \hline
2 &
  Rear-end &
  \begin{tabular}[c]{@{}l@{}}The front end of one vehicle collides \\ with the rear end of another vehicle, while \\ the two vehicles are traveling in the same \\ direction.\end{tabular} &
  \begin{tabular}[c]{@{}l@{}}'Same Direction Rear End',\\ 'Same Direction Rear End Left Turn',\\ 'Same Direction Rear End Right Turn'\end{tabular} &
  'Rear End' \\ \hline
3 &
  Angle &
  \begin{tabular}[c]{@{}l@{}}A crash where two motor vehicles \\ impact at an angle. For example, \\ the front of one motor vehicle \\ impacts the side of another motor \\ vehicle.\end{tabular} &
  \begin{tabular}[c]{@{}l@{}}'Same Movement Angle',\\ 'Angle Meets Right Turn',\\ 'Same Direction Right Turn',\\ 'Angle Meets Left Turn',\\ 'Same Direction Left Turn',\\ 'Same Direction Both Left Turn'\end{tabular} &
  \begin{tabular}[c]{@{}l@{}}'ANGLE (Front To Side)-\\ -(Other Than Left Turn)', \\ 'Left Turn', \\ 'Angle-\\ -Other Than Left Turn 2', \\ 'U Turn'\end{tabular} \\ \hline
4 &
  \begin{tabular}[c]{@{}l@{}}Sideswipe, \\ Same\\ Direction\end{tabular} &
  \begin{tabular}[c]{@{}l@{}}Two vehicles traveling in the\\ same direction impact one another \\ in a manner wherein the initial engagement\\ does not overlap the corner of either vehicle \\ so that there is no significant involvement \\ of the front or rear surface areas. \\ The impact then swipes along the surface \\ of the vehicle parallel to the direction of travel.\end{tabular} &
  `Same Direction Sideswipe' &
  `Sideswipe Same Direction' \\ \hline
5 &
  \begin{tabular}[c]{@{}l@{}}Sideswipe,\\ Opposite\\ Direction\end{tabular} &
  \begin{tabular}[c]{@{}l@{}}Two vehicles traveling in opposite directions \\ impact one another in a manner wherein the \\ initial engagement does not overlap the corner\\ of either vehicle so that there is no significant \\ involvement of the front or rear surface areas. \\ The impact then swipes along the surface of \\ the vehicle parallel to the direction of travel.\end{tabular} &
  \begin{tabular}[c]{@{}l@{}}'Opposite Direction Both Left Turn', \\ `Opposite Direction Sideswipe'\end{tabular} &
  `Sideswipe Opposite Direction' \\ \hline
6 &
  Head on &
  \begin{tabular}[c]{@{}l@{}}The front end of one vehicle impacts with \\ the front end of another vehicle, while the \\ two vehicles are traveling in opposite \\ directions.\end{tabular} &
  \begin{tabular}[c]{@{}l@{}}'Head On', \\ 'Head On Left Turn',   \\ `Angle Meets Left Turn Head On'\end{tabular} &
  `Head On' \\ \hline
7 &
  Rear to Rear &
  \begin{tabular}[c]{@{}l@{}}The rear end of a vehicle impacts with \\ the rear end of another. This can happen \\ when two vehicles are backing up.\end{tabular} &
   &
  `Rear To Rear' \\ \hline
8 &
  Front to Rear & \begin{tabular}[c]{@{}l@{}}The front end of one vehicle impacts with \\ the rear end of another vehicle, while the \\ two vehicles are traveling in the same \\ direction.\end{tabular}
   & 
   &
   \\ \hline
9 &
  Front to Front & \begin{tabular}[c]{@{}l@{}}The front end of one vehicle impacts with \\ the front end of another vehicle, while the \\ two vehicles are traveling in opposite \\ directions.\end{tabular}
   &
   &
   \\ \hline
10 &
  Rear to Side & \begin{tabular}[c]{@{}l@{}}The rear end of one vehicle impacts with \\ the side of another vehicle. This can occur \\ when one vehicle is moving forward and \\ another is attempting to merge or change \\ lanes, or when a vehicle backing out of a \\ parking space collides with a vehicle \\ moving  perpendicularly to it. \end{tabular}
   &
   &
  \begin{tabular}[c]{@{}l@{}}`Rear To Side', \\ '10'\end{tabular} \\ \hline
99 &
  Unknown &
  \begin{tabular}[c]{@{}l@{}}If this attribute is used, an explanation \\ in the narrative is recommended.\end{tabular} &
  \begin{tabular}[c]{@{}l@{}}`Other', \\ `Unknown', \\ `Not Applicable'\end{tabular} &
  \begin{tabular}[c]{@{}l@{}}`Other', \\ `Unknown', \\ NULL\end{tabular} \\ \hline
\end{tabular}%
}
\label{tab:Collisionmanner}
\end{table}


This pre-processing step ensures that the data is on a uniform basis for features and allows us to continue generating decision trees. To facilitate data analysis, categorical variables in the dataset are encoded as numbers for easy manipulation and calculation.

Overall, the unified feature engineering step is an important part of the data analysis process as it helps ensure data consistency, reliability, and accuracy in generating results. Furthermore, the unified feature engineering manner could provide suggestions for traffic accident data collection to facilitate accident recordings of different states maintaining unified. 

\subsection{Tree-based Models}
Going forward, 
we develop tree-based models that extract rule logic from decision trees and utilize robustness verification methods on XGBoost models. 

Our analysis 
involves using a set of independent variables that are commonly found in most states. These independent variables may differ between states and include factors such as weather conditions, lighting conditions, road surface conditions, collision type, causes of accidents, and the number of vehicles involved. Our aim is to classify accidents into two categories: those that result in fatalities, such as injuries or death, and minor accidents that do not result in any fatalities.

To accomplish our goal, we initially constructed several binary decision trees using a selected maximum depth and a minimum number of samples for the split. 
The maximum depth was chosen from [3,4,5],
and the minimum number of samples was chosen from [2,10,20,50].

Next, we aim to identify the best performance decision tree for each state.
All datasets are split into a training set (20\%) and a testing set (80\%), and we use grid search to find the best performance tree with the highest F1 score by running it on the test set. F1 score is the harmonic mean of precision and recall, and its value ranges between 0 and 1. A high F1 score indicates that the model has high precision and recall, while a low F1 score suggests that the model either lacks precision or is unable to identify all positive instances.
In the case of ties in the F1 score, we selected the tree with the lower depth since its effects are more interpretable. These chosen trees will be used in any further analysis.
Also, we consider evaluation metrics of accuracy, precision, and recall rate.
Accuracy is the percentage of correctly classified instances out of the total number of instances. 
Precision is the ratio of true positives (TP) to the total number of instances that the model predicted as positive. 
Recall rate is defined as the ratio of true positives (TP) to the sum of TP and false negatives (FN).

In Table~\ref{tab:detailed_analysis}, we provide additional information on the analysis for each state.

\begin{table}[h!]
\centering
\caption{Detailed information on the analysis for each state.}
\label{tab:detailed_analysis}
\resizebox{\columnwidth}{!}{%
\begin{tabular}{lllll}
\hline
{\textbf{State}}   & {\textbf{Input Features}} & {\textbf{\begin{tabular}[c]{@{}l@{}}Best DT\\ Performance\end{tabular}}}  & {\textbf{\begin{tabular}[c]{@{}l@{}}Best XGBoost\\ Performance\end{tabular}}}           & {\textbf{\begin{tabular}[c]{@{}l@{}}Human-understandable\\ Logic Involved Features\end{tabular}}}                                                                                                                                                                                               \\ \hline
{Arizona State}    & {\begin{tabular}[c]{@{}l@{}}1. junction relation\\ 2. collision manner\\ 3. light condition\\ 4. weather\\ 5. road surface condition\\ 6. driver violation manner one\\ 7. driver violation manner two\\ 8. driver action one\\ 9. driver action two\\ 10. alcohol use label\\ 11. drug use label\end{tabular}}                                                  & {\begin{tabular}[c]{@{}l@{}}Accuracy:  70.32\%\\ Precision:  56.43\%\\ Recall:  19.11\%\\ F1-Score:  28.56\%\end{tabular}} & {\begin{tabular}[c]{@{}l@{}}Accuracy: 71.74\%\\ Precision:  67.18\%\\ Recall:  17.47\%\\ F1-Score:  27.72\%\end{tabular}} & {\begin{tabular}[c]{@{}l@{}}1. driver violation manner one\\ 2. collision manner\\ 3. driver action one\\ 4. driver action two\\ 5. junction relation\\ 6. road surface condition\end{tabular}}                                                       \\ \hline
{Maryland State}   & {\begin{tabular}[c]{@{}l@{}}1. light description\\ 2. junction description\\ 3. collision type description\\ 4. surface condition description\\ 5. road condition description\\ 6. road division description\\ 7. fix object description\\ 8. weather description\\ 9. harm event description one\\ 10. harm event description two\\ 11. lane code\end{tabular}} & {\begin{tabular}[c]{@{}l@{}}Accuracy:  66\%\\ Precision:  56.4\%\\ Recall:  38.53\%\\ F1-Score:  45.78\%\end{tabular}}     & {\begin{tabular}[c]{@{}l@{}}Accuracy: 67.33\%\\ Precision:  60.34\%\\ Recall:  35.94\%\\ F1-Score:  45.05\%\end{tabular}} & {\begin{tabular}[c]{@{}l@{}}1. harm event description one\\ 2. collision type description\\ 3. fix object description\\ 4. junction description\\ 5. harm event description two\\ 6. weather description\\ 7. road division description\end{tabular}} \\ \hline
{New York State}   & {\begin{tabular}[c]{@{}l@{}}1. lighting conditions\\ 2. collision type descriptor\\ 3. road descriptor\\ 4. weather conditions\\ 5. traffic control device\\ 6. road surface conditions\\ 7. pedestrian bicyclist action\\ 8. event descriptor\\ 9. number of vehicles involved\end{tabular}}                                                                    & {\begin{tabular}[c]{@{}l@{}}Accuracy:  75.09\%\\ Precision:  84.51\%\\ Recall:  20.72\%\\ F1-Score:  33.28\%\end{tabular}} & {\begin{tabular}[c]{@{}l@{}}Accuracy: 75.55\%\\ Precision:  83.91\%\\ Recall:  22.84\%\\ F1-Score:  35.91\%\end{tabular}} & {\begin{tabular}[c]{@{}l@{}}1. pedestrian bicyclist action\\ 2. event descriptor\\ 3. number of vehicles involved\\ 4. road descriptor\\ 5. traffic control device\\ 6. collision type descriptor\end{tabular}}                                       \\ \hline
{Washington State} & {\begin{tabular}[c]{@{}l@{}}1. weather\\ 2. address type\\ 3. collision type\\ 4. junction type\\ 5. light condition\\ 6. road condition\\ 7. inattention label\\ 8. hit parked car label\\ 9. speeding label\\ 10. vehicle number\\ 11. pedestrian number\\ 12. pedal cyclist number\\ 13. person number\end{tabular}}                                          & {\begin{tabular}[c]{@{}l@{}}Accuracy:  77.78\%\\ Precision:  77.15\%\\ Recall:  28.9\%\\ F1-Score:  42.04\%\end{tabular}}  & {\begin{tabular}[c]{@{}l@{}}Accuracy: 77.84\%\\ Precision:  74.71\%\\ Recall:  31.06\%\\ F1-Score:  43.88\%\end{tabular}} & {\begin{tabular}[c]{@{}l@{}}1. pedestrian number\\ 2. collision type\\ 3. person number\\ 4. pedal cyclist number\\ 5. weather\\ 6. vehicle number\\ 7. light condition\\ 8. junction type\\ 9. road condition\end{tabular}}                          \\ \hline
\end{tabular}%
}
\end{table}

\subsection{Accident Rules}
We can simply extract several severe accident rules from the decision trees mentioned above. For instance, for New York State, we have three rules. Here, we note the classification label as $label$, and the value mappings are: accidents with injure or death: $label_{yes}$, accidents without injure or death: $label_{no}$. We note the feature of pedestrian bicyclist action as $pba$, and the value mappings are: crossing, no signal: $pba_0$, crossing, with signal: $pba_1$, getting on/off vehicle: $pba_2$, in the roadway: $pba_3$, not in the roadway: $pba_4$, unknown: $pba_5$. We note the event descriptor as $ed$ and the value mappings are: 
collision with bicyclist: $ed_0$, collision with fixed object: $ed_1$, 
collision with animal: $ed_2$, non-collision: $ed_3$, collision with motor vehicles: $ed_4$, collision with pedestrian: $ed_5$, collision with railroad train: $ed_6$. We note the number of vehicles involved as $vno$. We note traffic control device as $tcd$, and the value mappings are: flashing light: $tcd_0$, none: $tcd_1$, officer: 
$tcd_2$, railroad crossing: $tcd_3$, school zone: $tcd_4$, stop sign: 
$tcd_5$, 
no passing zone: $tcd_6$, traffic signal: $tcd_7$, unknown: $tcd_8$, work area: $tcd_9$.
\begin{enumerate}
    \item $((pba\implies pba_0) \lor (pba\implies pba_1) \lor (pba\implies pba_2) \lor (pba\implies pba_3) \lor (pba\implies pba_4) \implies (label \implies label_{yes})$
    \item $((pba\implies pba_5) \land (ed\implies ed_0) \implies (label \implies label_{yes})$
    \item $((pba\implies pba_5) \land ((ed\implies ed_1) \lor (ed\implies ed_2) \lor (ed\implies ed_3) \lor (ed\implies ed_4) \lor (ed\implies ed_5) \lor (ed\implies ed_6)) \land (vno \geq 2) \land((tcd \implies tcd_2) \lor (tcd \implies tcd_3) \lor (tcd \implies tcd_4) \lor (tcd \implies tcd_5) \lor (tcd \implies tcd_6) \lor (tcd \implies tcd_7) \lor (tcd \implies tcd_8) \lor (tcd \implies tcd_9)) \implies (label \implies label_{yes})$
\end{enumerate}

Thus, \textbf{Rule 1} and \textbf{Rule 2} indicate that if there are some specific pedestrian or bicyclist actions involved in accidents, we may imply these accidents are injurious or even fatal. The potential reason for this logic may relate to the vulnerability and lack of physical protection of pedestrians or bicyclists in running traffic. \textbf{Rule 3} indicates that although there's no pedestrian or bicyclist involved in the accidents, chaos situations above average level, which include more vehicles (more than two vehicles involved) and non-strong supervised traffic control (not supervised by a police officer), may also be implied as injurious or even fatal accidents.

\subsection{Robustness Verification}
In the context of machine learning, robustness verification typically involves testing the performance of a trained model against various perturbations of its input data, such as random noise or deliberate modifications designed to cause the model to make incorrect predictions. The goal of this process is to identify any weaknesses or vulnerabilities in the model's performance and to ensure that it can effectively handle unexpected inputs or situations.

Robustness verification is important because machine learning models are often used in high-stakes applications where incorrect predictions can have serious consequences, such as in medical diagnosis or autonomous driving. By verifying the robustness of these models, we can increase their reliability and safety, and reduce the risk of errors or failures.

For decision tree or decision tree ensembles, formal robustness verification involves finding the exact minimal adversarial perturbation or a guaranteed lower bound of it. Here, we give the definition of \textbf{minimal adversarial perturbation} in \eqref{equation1}. For the input sample x, assuming that $y_0 = f(x)$ is the correct label, where $f(.)$ mean the tree model, and if we add $\delta$ to x could change the prediction for sample x, the minimal $\delta$ is the minimal adversarial perturbation, noted as $r^*$:
\begin{equation}
r^* = \min_\delta \|\delta\|_\infty 
\qquad s.t. \ f(x+\delta) \neq y_0
\label{equation1}
\end{equation}

For a single tree, a given sample $x = [x_1,...,x_d]$ with $d$ dimensions will start from the root node and traverse the tree to reach a final leaf node based on the decision threshold of each decision node. For example, for decision node $i$, which has the two children (the left child and the right child), if the samples will separate based on feature $t_i$ and the threshold value is $\eta_i$, x will be passed to the left child if $x_{t_i} \leq \eta_i$ and to the right child otherwise.
The main idea of the single tree verification is to compute a $d$-dimensional box for each leaf node such that any sample in this box will fall into this leaf. Mathematically, the node $i$'s box is defined as the Cartesian product $B^i =(l_1^i,r_1^i] \times \dots \times (l_d^i,r_d^i]$. 
More specifically, if p, q are node $i$’s left and right child node respectively, then we can set their boxes $B^p=(l_1^p,r_1^p] \times \dots \times (l_d^p,r_d^p]$ and $B^q =(l_1^q,r_1^q] \times \dots \times (l_d^q,r_d^q]$ by setting \eqref{equation2}:

\begin{equation}
\begin{split}
    (l_t^p,r_t^p] =
\begin{cases}
(l_t^i,r_t^i],  & \text{if $t \neq t_i$} \\
(l_t^i,\min \{ r_t^i, \eta_i \}], & \text{if $t = t_i$}
\end{cases} \\
(l_t^q,r_t^q] =
\begin{cases}
(l_t^i,r_t^i],  & \text{if $t \neq t_i$} \\
(\max \{l_t^i, \eta_i \}, r_t^i], & \text{if $t = t_i$}
\end{cases}
\label{equation2}
\end{split}
\end{equation}

With the boxes computed for each leaf node, the minimum perturbation required to change x to go to a leaf node $i$ can be written as a vector $\epsilon(x,B^i) \in \mathbb{R^d}$ defined as \eqref{equation3}:

\begin{equation}
\epsilon(x,B^i)_t =
\begin{cases}
0,  & \text{if $x_t \in (l_t^i,r_t^i]$} \\
x_t - r_t^i,  & \text{if $x_t > r_t^i$} \\
l_t^i - x_t, & \text{if $x_t \leq l_t^i$}
\end{cases}
\label{equation3}
\end{equation}

Thus, the minimal adversarial perturbation could be computed as $r^* = \min_{i:v_i \neq y_0} \| \epsilon(x,B^i) \|_\infty$.

\section{Evaluations}

\subsection{Data}
When searching for ``crash'' datasets, 72 related links are displayed on Data.gov. By categorizing the data by state, we can see that there are 1 federal-level dataset and 10 state-level datasets, which include Arizona,
Louisiana,
Iowa,
Maryland, 
Massachusetts,
New York,
North Carolina,
Pennsylvania, 
Tennessee,
and Washington.
These datasets are further divided into state-level, city-level, and county-level datasets. To make the data more manageable, we select 
Arizona,
Maryland,
New York,
and Washington. We chose these states based on the size and format of the data.

\subsection{Evaluation of Decision Trees}
\begin{figure}[ht]
    \centering
    \includegraphics[width=.85\textwidth]{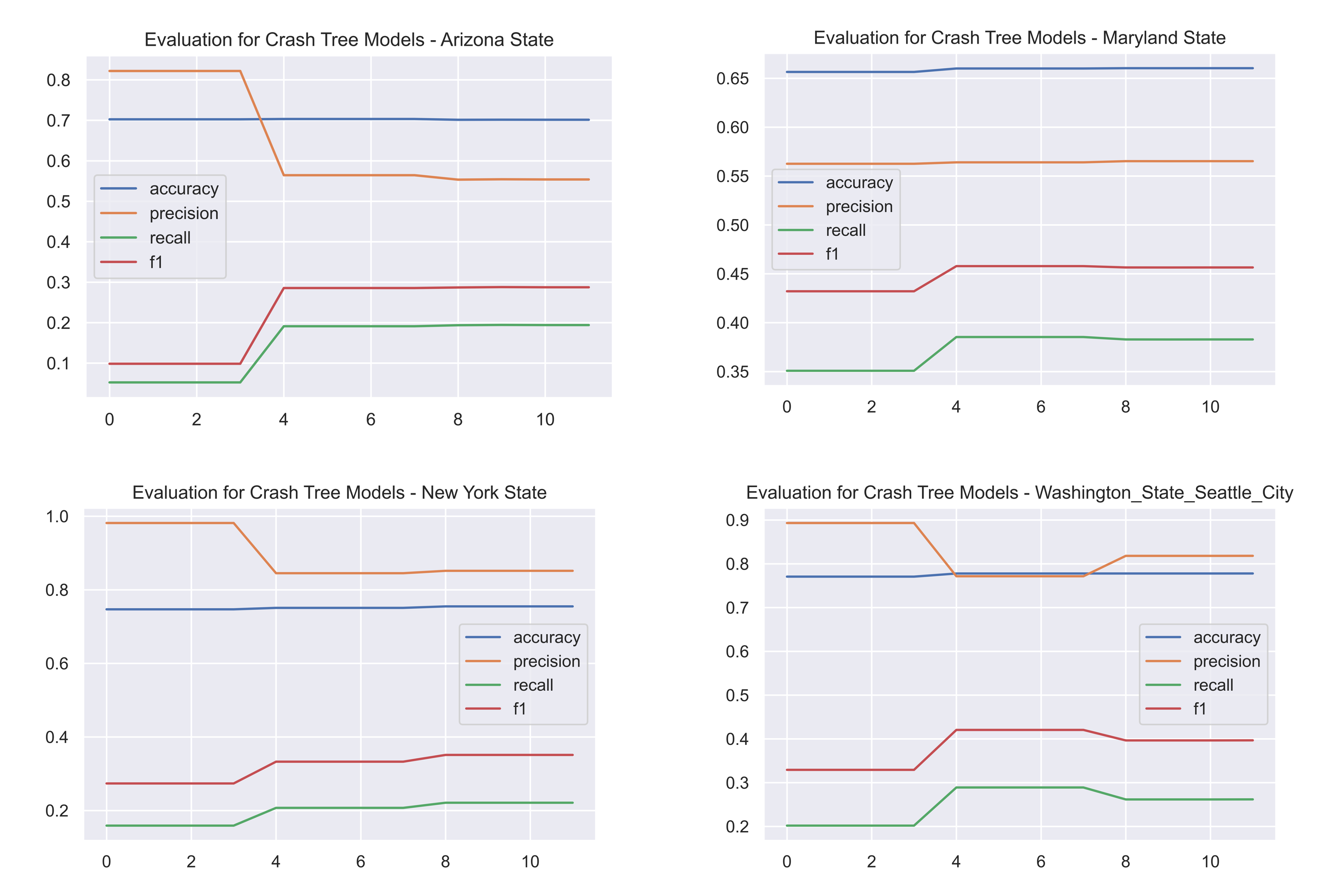}
    \caption{Evaluation of Decision Trees of the Chosen States}
    \label{fig:evaluation_DT}
\end{figure}

In this study, we fine-tuned our decision tree models for each state, focusing on optimizing performance metrics such as accuracy, precision, recall, and F1 score. To achieve this, we explored various hyperparameters as aforementioned.
By closely monitoring the impact of these adjustments on the evaluation metrics, we were able to identify the best-performing model for each state. The evaluation results are shown in Figure \ref{fig:evaluation_DT}.

\subsection{Evaluation of Robustness Verification}
In this study, we utilize a state-of-the-art decision tree robustness verification method\cite{chen2019robustness} to evaluate the robustness of our decision trees under adversarial attacks.
Here we don't solve the problem to get the exact $r^*$, but to find a \emph{lower bound} \underline{r}, which guarantees that no adversarial sample exists within radius \underline{r} because the exact verification problem is NP-complete. A high-quality lower bound \underline{r} should be close to $r^*$. The ``average bound'' is the lower bound of minimum adversarial distortion averaged over all test examples. A larger value typically indicates better overall robustness of the model.
Also, the ``verified error'' is the upper bound of error under any attacks, which could be treated as an index to measure the \emph{worst case} performance.

By default, the model evaluates the robustness under adversarial attacks of a fixed number (1000) of test points from a given dataset (in LIBSVM format), using an initial epsilon value for the binary search process. We consider 5 parameters:
\begin{enumerate}
    \item \textbf{eps\_init}: the first epsilon in the binary search. This epsilon is also used to compute verified errors.
    \item \textbf{max\_search}: maximum number of binary searches for searching the largest epsilon that our algorithm can verify. By setting max\_search to 1, the algorithm will disable binary search and only return the verified error at a certain epsilon.
    \item \textbf{max\_level}:  maximum number of levels of clique search. A larger number will produce better quality bounds but the verification process becomes much slower. 
    \item \textbf{max\_clique}: maximum number of nodes in a clique.
    \item \textbf{dp}: by setting DP to 1, the algorithm will use dynamic programming, to sum up nodes on the last level. The default is 0, which means DP is not used, and a simple summation is used instead.
\end{enumerate}

For a more in-depth analysis, we run this verification model with different configurations against all our decision trees, exploring the impact of various parameters on the model's ability to verify robustness. The verification results are shown in Table \ref{tab:robust}.
Here, we gain two observations: first, since the average bound larger is better and the verified error smaller is better, we see the tree models of Maryland State and Arizona State perform better in robustness than others, which is partly because the two states have larger datasets and the data quality may be better too. Second, we can compare the results under setting 1 and each of setting 2-5, since only one parameter is changed in the latter settings compared to the settings. We can see that (1) the initial epsilon almost doesn't have an influence on the results; (2) binary search is crucial to find the average bound; (3) the larger the maximum number of levels of clique search, the larger maximum number of nodes in a clique and dynamic programming is beneficial for locating a better robustness estimation.

\begin{table}[t]
\caption{Robustness verification evaluations under different experimental settings ($\uparrow$ represents that larger values are deemed more favorable or superior within the given context, Conversely, $\downarrow$ represents that smaller values are considered more advantageous or preferable within the scope of the situation).}
\resizebox{\columnwidth}{!}{%
\begin{tabular}{|c|cc|cc|cc|}
\hline
experiment   settings &
  \multicolumn{2}{c|}{\begin{tabular}[c]{@{}c@{}} \underline{setting 1:} \\ eps\_init:       0.3, \\      max\_clique:   2,\\       max\_search:   10,\\      max\_level:    1\end{tabular}} &
  \multicolumn{2}{c|}{\begin{tabular}[c]{@{}c@{}} \underline{setting 2:} \\ \textbf{eps\_init:       0.5}, \\      max\_clique:   2,\\       max\_search:   10,\\      max\_level:    1\end{tabular}} &
  \multicolumn{2}{c|}{\begin{tabular}[c]{@{}c@{}} \underline{setting 3:} \\ eps\_init:     0.3, \\      max\_clique:   2,\\       \textbf{max\_search:   1},\\      max\_level:    1\end{tabular}} \\ \hline
\textbf{state} &
  \textbf{avg. bound} $\uparrow$ &
  \textbf{verified error} $\downarrow$ &
  \textbf{avg. bound} $\uparrow$ &
  \textbf{verified error} $\downarrow$ &
  \textbf{avg. bound} $\uparrow$ &
  \textbf{verified error} $\downarrow$ \\ \hline
NY &
  0.23 &
  0.77 &
  0.23 &
  0.77 &
  0.07 &
  0.77 \\
ML &
  \textbf{0.59} &
  \textbf{0.40} &
  \textbf{0.59} &
  \textbf{0.41} &
  \textbf{0.18} &
  \textbf{0.40} \\
WA &
  0.56 &
  0.44 &
  0.56 &
  0.44 &
  0.17 &
  0.44 \\
AZ &
  0.51 &
  0.49 &
  0.51 &
  0.49 &
  0.15 &
  0.49 \\ \hline
experiment settings &
  \multicolumn{2}{c|}{\begin{tabular}[c]{@{}c@{}} \underline{setting 4:} \\ eps\_init:       0.3, \\      max\_clique:   2,\\       max\_search:   10,\\      \textbf{max\_level:    2}\end{tabular}} &
  \multicolumn{2}{c|}{\begin{tabular}[c]{@{}c@{}}  \underline{setting 5:} \\ eps\_init:       0.3, \\      \textbf{max\_clique:   4},\\       max\_search:   10,\\      max\_level:    1\end{tabular}} &
  \multicolumn{2}{c|}{\begin{tabular}[c]{@{}c@{}}  \underline{setting 6:} \\ eps\_init:     0.3, \\      max\_clique:   2,\\       max\_search:   10,\\      max\_level:    1,\\      \textbf{dp:    1}\end{tabular}} \\ \hline
\textbf{state} &
  \textbf{avg. bound} $\uparrow$ &
  \textbf{verified error} $\downarrow$ &
  \textbf{avg. bound} $\uparrow$ &
  \textbf{verified error} $\downarrow$ &
  \textbf{avg. bound} $\uparrow$ &
  \textbf{verified error} $\downarrow$ \\ \hline
NY &
  0.51 &
  0.49 &
  0.51 &
  0.49 &
  0.68 &
  0.32 \\
ML &
  0.64 &
  0.36 &
  0.64 &
  0.36 &
  0.64 &
  0.35 \\
WA &
  0.62 &
  0.36 &
  0.62 &
  0.38 &
  0.67 &
  0.33 \\
AZ &
  \textbf{0.68} &
  \textbf{0.32} &
  \textbf{0.68} &
  \textbf{0.32} &
  \textbf{0.71} &
  \textbf{0.29} \\ \hline
\end{tabular}%
}
\label{tab:robust}
\end{table}

\section{Discussion and Conclusion}


The application of robustness verification to tree-based models assumes paramount significance, particularly within transportation systems. This emphasis stems from the indispensable need for accurate and dependable predictions. Our meticulous approach to validation serves as a robust safeguard, ensuring the reliability and safety of these models in critical contexts.

In this work, we compile a comprehensive real-world dataset encompassing four states within the United States. This dataset serves as the foundation for training decision tree models dedicated to predicting severe accidents. Through this process, we extract invaluable accident detection insights and rule-based logic. Notably, we introduce an innovative approach to constructing tree-based models, tailored for the identification of high-risk driving scenarios. Subsequently, we employ robustness verification techniques on these tree ensembles, a pivotal stride that gauges both our confidence level and the limitations that safeguard the logic against potential failures.

In conclusion, we have determined that the tree path rules possess both meaningful and explicable qualities. Moreover, the implementation of a unified feature engineering process holds the potential to foster a more standardized and uniform paradigm for the collection of traffic accident data in each state. This, in turn, would facilitate the amalgamation of available data nationwide, resulting in a more consistent and comprehensive dataset. Additionally, the utilization of a larger dataset accompanied by enhanced recording quality has the potential to correspondingly elevate the level of robustness. Lastly, it is noteworthy that an increased maximum number of levels in clique search, along with a larger maximum number of nodes in a clique and dynamic programming, holds the potential to significantly enhance the accuracy of robustness estimation.



\section*{Acknowledgment}
This material is based upon work supported by the National Science Foundation under Grant 2151500.


\end{document}